\documentclass[11pt]{article}
\usepackage[a4paper]{geometry}
\usepackage{clic2017}
\usepackage{times}
\usepackage{url}
\usepackage{latexsym}
\usepackage{graphicx}
\usepackage{amsmath}
\usepackage{titlecaps}
\usepackage{multicol}
\usepackage{ctable}
\usepackage[T1]{fontenc}
\usepackage{enumitem}
\newcommand{\argmax}{\arg\!\max}
  \title{Analysis of Italian Word Embeddings}

\author{Rocco Tripodi \\
  Ca' Foscari University of Venice \\
  {\tt rocco.tripodi@unive.it} \\\And
  Stefano Li Pira \\
  University of Warwick \\
  {\tt stefano.li-pira@wbs.ac.uk} \\}

\date{}

\begin{document}
\maketitle
\begin{abstract}
  \textbf{English.}  In this work we analyze the performances of two of the most used word embeddings algorithms, skip-gram and continuous bag of words on Italian language. These algorithms have many hyper-parameter that have to be carefully tuned in order to obtain accurate word representation in vectorial space. We provide an extensive analysis and an evaluation, showing what are the best configuration of parameters for specific analogy tasks.
	\end{abstract}

\begin{abstract-alt}
 \textrm{\bf{Italiano.}} In questo lavoro analizziamo le performances di due tra i pi\`{u} usati algoritmi di word embedding: skip-gram e continuous bag of words. Questi algoritmi hanno diversi iperparametri che devono essere impostati accuratamente per ottenere delle rappresentazioni accurate delle parole all'interno di spazi vettoriali. Presentiamo un'analisi accurata e una valutazione dei due algoritmi mostrando quali sono le configurazioni migliori di parametri su specifiche applicazioni.
\end{abstract-alt}

\section{Introduction}
The distributional hypothesis of language, set forth by \newcite{firth1935technique} and \newcite{harris1954distributional}, states that the meaning of a word can be inferred from the contexts in which it is used. Using the co-occurrence of words in a large corpus, we can observe for example that the contexts in which \emph{client} is used are very similar to those in which \emph{customer} occur, while less similar to those in which \emph{waitress} or \emph{retailer} occur. A wide range of algorithms have been developed to exploit these properties. Recently, one of the most widely used method in many natural language processing (NLP) tasks is word embeddings \cite{bengio2003neural,mikolov2010recurrent,mikolov2013efficient}. It is based on neural network techniques and has demonstrated  to capture semantic and syntactic properties of words taking as input raw texts without other sources of information. It represents each word as a vector such that words that appear in similar contexts are represented with similar vectors \cite{collobert2008unified,mikolov2013efficient}. The dimensions of the word are not easily interpretable and, with respect to explicit representation, they do not correspond to specific concepts.

In \newcite{mikolov2013efficient}, the authors propose two different models that seek to maximize, respectively, the probability of a word given its context (Continuous bag-of-word model), and the probability of the surrounding words (before and after the current word) given the current word
(Skip-gram model). In this work we seek to further explore the relationships by generating word embedding for over 40 different parameterizations of the continuous bag-of-words (CBOW) and the skip-gram (SG) architectures, since as shown in \newcite{levy2015improving} the choice of the hyper-parameters heavily affect the construction of the embedding spaces.

%In this work we seek to explore the relationships by generating word embedding for over 40 different parameterizations of the continuous bag-of-words (CBOW) and the skip-gram (SG) architectures, since as shown in \newcite{levy2015improving} the choice of the hyper-parameters heavily affect the construction of the embedding spaces. The main difference among the two architectures is that CBOW tries to predict a word given its context while SG tries to predict the context given a word.

Specifically our contributions include:
\begin{itemize}[leftmargin=*]\setlength\itemsep{0.3em}
\item \textbf{Word embedding.}  The analysis of how different hyper-parameters can achieve different accuracy levels in relation recovery tasks \cite{mikolov2013efficient}. 
\item \textbf{Morpho-syntactic and semantic analysis.} Word embeddings have demonstrated to capture semantic and syntactic properties, we compare two different objectives to recover relational similarities for semantic and morph-syntactical tasks. 
\item \textbf{Qualitative analysis.} We investigate problematic cases.
\end{itemize}

%In this work, we generate the word embedding for the Italian language. We tried to capture how the model is able to represent syntactical and semantical relationships in the Italian language given its higher morphological complexity with respect to English.

%\section{Related works}
%In \cite{berardi2015word}, the SG parameters were chosen following the best performances of the algorithm on English. This choice is arbitrary and does not take into account that English and Italian are very different at grammatical and syntactical level. In fact, Italian is an inflected language while English allows less variation in word order. Furthermore, the verb conjugation in English retains the same word form, while in Italian it changes for each person, number, tense and mood. For these reasons we decide to investigate the influence of the hyper-parameters on the construction of word spaces.

\section{Related works}
The interest that word embedding models have achieved in the NLP international community has recently been confirmed by the increasing number of studies that are adopting these algorithms in languages different from English. One of the first example is the Polyglot project that produced word embedding for 117 languages \cite{al2013polyglot}. They demonstrated the utility of word embedding, achieving, in a part of speech tagging task, performances competitive with the state-of-the art methods in English. \newcite{attardi2014adapting} have done the first attempt to introduce word embedding in Italian obtaining similar results. They have shown that, using word embedding, they obtained one of the best accuracy levels in a named entity recognition task.  

However, these optimistic results are not confirmed by more recent studies. Indeed the performance of word embedding are not directly comparable in the accuracy test to those obtained in the English language. For example, \newcite{attardi2014dependency} combining the word embeddings in a dependency parser have not observed improvements over a baseline system not using such features. \newcite{berardi2015word} found a $~ 47\%$ accuracy on the Italian versus $~ 60\%$ accuracy on the English. The results may be a sign of a higher complexity of Italian with respect to English as we will see section \ref{sec:exp}.

Similarly, recent work that trained word embeddings on tweets have highlighted some criticalities. One of these aspects is how the morphology of a word is opaque to word embeddings. Indeed, the relatedness of the meaning of a lemma's different word forms, its different string representations, is not systematically encoded. This means that in morphologically rich languages with long-tailed frequency distributions, even some word embedding representations for word forms of common lemmata may become very poor \cite{kim2016character}.

For this reason, some recent contribution on Italian tweets have tried to capture these aspects. \newcite{tamburini2016bilstm} trained SG on a set of 200 million tweets. He proposed a PoS-tagging system integrating neural representation models and a morphological analyzer, exhibiting a very good accuracy. Similarly, \newcite{stemle2016bot} proposes a system that uses word embeddings and augment the WE representations with character-level representations of the word beginnings and endings.

We have observed that in these studies the authors used either the most common set-up of parameters gathered from the literature \cite{tamburini2016bilstm,stemle2016bot,berardi2015word} or an arbitrary number  \cite{attardi2014dependency,attardi2016convolutional}. Despite the relevance given to these parameters in the literature \cite{goldberg2017neural} we have not seen studies that analyze the different strategies behind the possible parametrization. In the next section, we propose a model to deepen these aspects.

\begin{table}[]
\centering
\resizebox{0.4\textwidth}{!}{%
\begin{tabular}{l|l|l} \specialrule{.05em}{.05em}{.05em} 
\textbf{HP} & \textbf{SG} & \textbf{CBOW} \\ \hline
$dim$ & $200$, $300$, $400$, $500$ & $200$, $300$, $400$, $500$ \\
$w$ & $3$, $5$ & $2$, $5$ \\
$m$ & $1$, $5$ & $1$, $5$\\
$n$ & $1$, $5$, $10$ & $1$, $5$, $15$ \\ \specialrule{.1em}{.05em}{.05em} 
\end{tabular}}\caption{Hyper-parameters}\label{tab:hpar}
\end{table}

\section{Italian word embeddings}
Previous results on the word analogy tasks have been reported using vectors obtained with proprietary corpora \cite{berardi2015word}. To make the experiments reproducible, we trained our models on a dump of the Italian Wikipedia (dated 2017.05.01), from which we used only the body text of each articles. The obtained texts have been lowercased and filtered according to the corresponding parameter of each model. The corpus consists of 994.949 sentences that result in 470.400.914 tokens.

The hyper-parameters used to construct the different embeddings for the SG and the CBOW models are: the size of the vectors ($dim$), the window size of the words contexts ($w$), the minimum number word occurrences ($m$) and the number of negative samples ($n$). The values that these hyper-parameters can take are shown in Table \ref{tab:hpar}.

\begin{table}[]
\centering
\resizebox{0.485\textwidth}{!}{%
\begin{tabular}{l|l} \specialrule{.1em}{.05em}{.05em} 
\textbf{Morphosyntactic} & \textbf{Semantic} \\ \hline
 adjective-to-adverb & capital-common-countries \\
 opposite & capital-world \\
 comparative & currency \\
 superlative (assoluto) & city-in-state \\
 present-participle (gerundio) & regione capoluogo \\
 nationality-adjective &  \\
 past-tense &  \\
 plural &  \\
 plural-verbs (3rd person) &  \\
 plural-verbs (1st person) &  \\
 remote-past-verbs (1st person) &  \\
 noun-masculine-feminine-singular &  \\
 noun-masculine-feminine-plural & \\ \hline
 \#10.876 & \#8.915 \\ \specialrule{.1em}{.05em}{.05em} 
\end{tabular}}\caption{Relation types}\label{tab:reltyp}
\end{table}

\section{Evaluation}
The obtained embedding\footnote{The trained vectors with the best performances are available at http://roccotripodi.com/ita-we} spaces are evaluated on an \emph{word analogy} task, using a enriched version of the Google word analogy test \cite{mikolov2013efficient}, translated in Italian by \cite{berardi2015word}. 
%\footnote{The dataset is available at http:// together with the list of corrections that we made on it}.
It contains 19.791 questions and covers 19 relations types. 6 of them are semantic and 13 morphosyntactic (see Table \ref{tab:reltyp}). The proportions of these two types of question is balanced as shown in Table \ref{tab:reltyp}. 

To recover these relations two different methods are used: \textsc{3CosAdd} (Eq. \ref{eq:cosadd}) \cite{mikolov2013efficient} and \textsc{3CosMul} (Eq. \ref{eq:cosmul}) \cite{levy2014linguistic} to compute vectors analogies:

\begin{equation}\label{eq:cosadd}
\text{\textsc{3CosAdd} } \argmax_{b^*\in V}{cos(b^*, b - a + a^*)}
\end{equation}

\begin{equation}\label{eq:cosmul}
\text{\textsc{3CosMul} } \argmax_{b^*\in V}{\frac{cos(b^*,b)cos(b^*,a^*)}{cos(b^*,a)+\epsilon}}
\end{equation}

\noindent These two measures try to capture different relations between word vectors. The idea behind these measures is to use the cosine similarity to recover the vector of the hidden word ($b^*$) that has to be the most similar vector given two positive and one negative word. In this way, it is possible to model relations such as \emph{queen} is to \emph{king} what \emph{woman} is to \emph{man}. In this case, the word \emph{queen} ($b^*$) is represented by a vector that has to be similar to \emph{king} ($b$) and \emph{woman} ($a^*$) and different to \emph{man} ($a$). The two analogy measures slightly differ in how they weight each aspect of the similarity relation. \textsc{3CosAdd} allows one sufficiently large term to dominate the expression \cite{levy2014linguistic}, \textsc{3CosMul} achieves a better balance amplifying the small differences between terms and reducing the larger ones \cite{levy2014linguistic}. As explained in \newcite{levy2014linguistic}, we expect \textsc{3CosMul} to over-perform \textsc{3CosAdd} in evaluating both the syntactic and the semantic tasks as it tries to normalize the strength of the relationships that the hidden term has both with the attractor terms and with the repellers term. 

\begin{figure*}[h]
\includegraphics[width=1\textwidth,trim={5.15cm 0 4.25cm 0},clip]{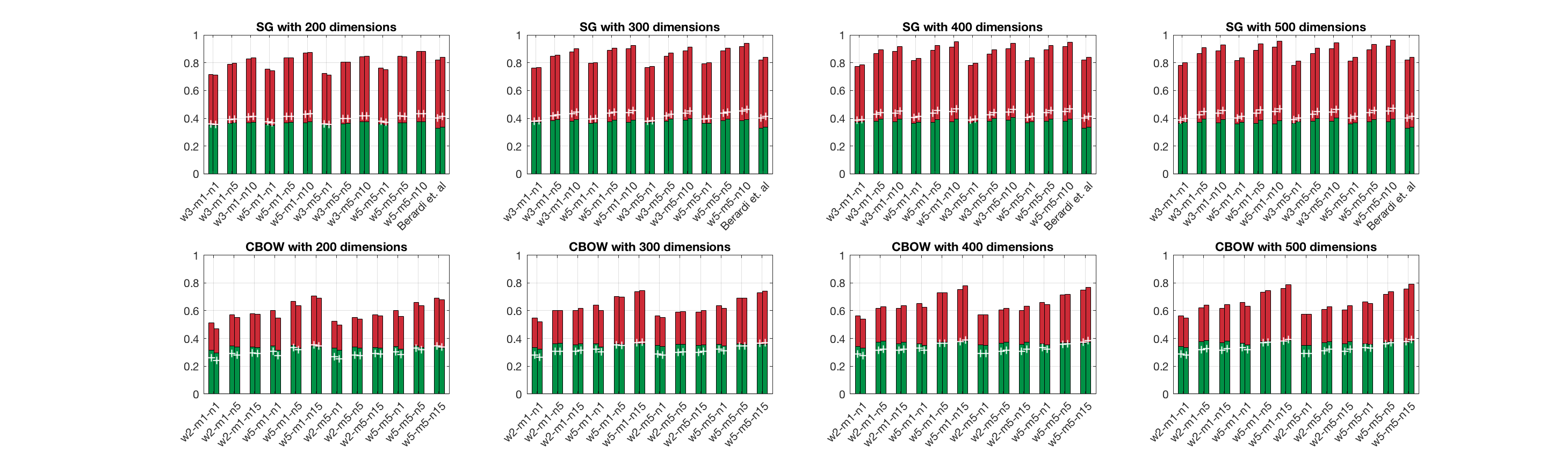}\caption{Results as accuracy with different hyper-parameters ($y$ axis) using the \textsc{3CosAdd} (left bar) and the \textsc{3CosMul} (right bar) formula. The green part of the bars  indicates the accuracy on the morpho-syntactic task whereas the red one the accuracy on the semantic task. The $+$ sign on each bar indicates the accuracy on the entire dataset. The upper row of the figure shows the results of the SG algorithm and the bottom row the results of CBOW. The last two bars of the SG plots indicates the results obtained using the vectors made available by \cite{berardi2015word}}\label{fig:res}
\end{figure*}

\subsection{Experimental results}\label{sec:exp}
The results of our evaluation are presented in Figure \ref{fig:res}. The main trend that it is possible to notice is that accuracy increases as the number of dimensions of the embedded vectors increases. This indicates that Italian language benefits of a rich representation that can account for its rich morphology. Another important trend that emerges is the fact that the parameters have the same effect on both algorithms and that they perform very differently on all the tasks. CBOW has very low accuracy compared to SG. We can also see that the $dim$ hyper-parameter is not correlated with the dimension of the vocabulary (model complexity) as one should expect. In fact, with increasing values of $dim$ the accuracy increases whatever is the value of $m$. This hyper-parameter heavily affects the vocabulary length (see Table \ref{tab:voc}). However the $dim$ hyper-parameter seems to be correlated only with the accuracy in the semantic tasks while the performances on the morpho-syntactic tasks seems not to have a big bust increasing the dimensionality.

%An interesting aspect to notice is the fact that a large window ($w=5$), a reduced vocabulary (m=5) and a many negative examples ($n=10$) have a beneficial effect on the semantic tasks

%In the next sub-section we will provide further analysis of these errors in order to understand why dimensionality is not a critical component in increasing the accuracy of the syntactical relationships.

With respect to the size of the context ($w$) used to create the words representations we do not observe a clear difference between the 18 pairs both in the SG and in the CBOW. On the contrary a clear trend can be observed varying the $n$ hyper-parameter, with $n=1$  the accuracy was significantly lower than the one we obtained with $n=5$ or $n=10$. Increasing the number of negative samples constantly increases the accuracy. 

These results support also the claim put forward by \cite{levy2014linguistic} that the \textsc{3CosMul} method is more suited to recover analogy relations. In fact, we can see that on average the right bars of the plots are higher than the left.

%There is some intuition behind ignoring negative values. Positive associations (e.g., \textit{Milano} and \textit{tram}) are more likely to be find while it is much harder to provide negative ones (\textit{Milano} and \textit{gondola}). This suggests that the perceived similarity of two words is influenced more by the positive context they share than by the negative context they share. Therefore it makes some intuitive sense to associate a loss to the negatively associated contexts rather then just considering them poorly informative. A notable exception seems to be the case of syntactic similarity. Indeed as noted in \newcite{goldberg2017neural} "all verbs share a very strong negative association with being preceded by determiners, and past tense verbs have a very strong negative association to be preceded by \textit{be} verbs and modals".

\begin{table}[]
\centering
\resizebox{0.25\textwidth}{!}{%
\label{}
\begin{tabular}{l|l|l} \specialrule{.1em}{.05em}{.05em} 
\textbf{m=1} & \textbf{m=5} & \textbf{Berardi} \\\hline
 3.227.282 & 847.355 & 733.392 \\ \specialrule{.1em}{.05em}{.05em} 
\end{tabular}}\caption{Vocabulary length}\label{tab:voc}
\end{table}

\subsection{Error analysis}
If we restrict the error analysis to the most macroscopic differences in figure \ref{fig:res} we can compare three different parametrizations: SG-200 w5-m5-n1, SG-500 w5-m5-n1, SG-500 w5-m5-n10. In this way we can analyze the results obtained changing the number of dimensions of the vectors and the role played by $n$. In Table \ref{tab:err} the total number of errors and the number of different words that have not been recovered by each parametrization are presented. 
\begin{table}[]
\centering
\resizebox{0.3\textwidth}{!}{%
\label{}
\begin{tabular}{l|l l} \specialrule{.1em}{.05em}{.05em} 
\textbf{Parametrization} & \textbf{\#errors} & \textbf{\#words} \\\hline
\textbf{SG-200-w5-m5-n10} & 10.113 & 543 \\ %\specialrule{.1em}{.05em}{.05em} 
\textbf{SG-500 w5-m5-n1} & 10.506 & 535 \\ %\specialrule{.1em}{.05em}{.05em} 
\textbf{SG-500 w5-m5-n10} & 9.337 & 525 \\ \specialrule{.1em}{.05em}{.05em} 
\end{tabular}}\caption{Total number of errors and number of different words that have not been recovered }\label{tab:err}
\end{table}
\noindent From this table we can see that most of the errors are done one a relatively small set of words. This phenomenon can be studied analyzing the  most problematic cases. In Table \ref{tab:mproblem} we can see the list of the most common errors ranked by frequency for each method.
\begin{table}[htb]
\centering
\resizebox{0.49\textwidth}{!}{%
\begin{tabular}{l c | l c | l c}\specialrule{.1em}{.05em}{.05em} 
SG-200-w5-m5-n10 & \# & SG-500 w5-m5-n1 & \# & SG-500 w5-m5-n10 & \# \\\hline
 california & 328 & california & 349 & california & 287 \\
 texas & 223 & texas & 224 & texas & 165 \\
 arizona & 164 & arizona & 164 & arizona & 145 \\
 florida & 144 & ohio & 142 & florida & 124 \\
 ohio & 135 & florida & 140 & ohio & 112 \\\specialrule{.1em}{.05em}{.05em} 
 %washington & 113 & washington & 114 & washington & 111 \\
 %nevada & 77 & nevada & 77 & nevada & 76 \\
 %georgia & 74 & georgia & 77 & georgia & 71 \\
 %pennsylvania & 73 & pennsylvania & 69 & pennsylvania & 62 \\
 %kentucky & 59 & kentucky & 67 & corona & 58
\end{tabular}}\caption{Most common errors}\label{tab:mproblem}
\end{table}
\noindent As we can see from these lists the errors are done on the same words and this because they are the most common in the dataset (e.g.: in the dataset there are 217 queries that require \textit{Florida} as answer compared to the 55 of \textit{Italia}). However if we compare the  frequency of these errors in the analogy test within the three parametrisation we can observe an improvement of approximately 15\% in accuracy with  SG-500 w5-m5-n10. Indeed, despite many errors are not recovered for any of the parametrisation, we can observe that approximately 21\% of the errors are recovered under certain parametrizations (Table  \ref{tab:solvederrors}).
%
%\begin{table}[htb]
%\centering
%\resizebox{0.35\textwidth}{!}{%
%\label{}
%\begin{tabular}{l|c } \specialrule{.1em}{.05em}{.05em}
%\textbf{Parametrization} & \textbf{\#errors solved}  \\\hline
%dim = 500 \& n = 10 & 873 \\ 
%\textit{solo} dim = 500 & 645 \\
%\textit{solo} n = 10 & 927  \\ \specialrule{.1em}{.05em}{.05em}
%\end{tabular}}\caption{Frequencies of errors solved. }\label{tab:solvederrors}
%\end{table}
%
\begin{table*}[htb]
\parbox[t]{.24\linewidth}{
%\begin{table}[htb]
\resizebox{0.24\textwidth}{!}{%
\begin{tabular}[]{l|c } \specialrule{.1em}{.05em}{.05em}
\textbf{Parametrization} & \textbf{\#errors solved}  \\\hline
dim = 500 \& n = 10 & 873 \\ 
\textit{solo} dim = 500 & 645 \\
\textit{solo} n = 10 & 927  \\ \specialrule{.1em}{.05em}{.05em}
\end{tabular}}\caption{Solved errors }\label{tab:solvederrors}
}
%\end{table}
\hfill
\parbox[t]{.73\linewidth}{
\resizebox{0.73\textwidth}{!}{%
\begin{tabular}{l | l | l}\specialrule{.1em}{.05em}{.05em} 
\textbf{dim = 500 \& n = 10}	 & 	\textbf{\textit{solo} n = 10} 	 & 	\textbf{\textit{solo} dim = 500} 	 \\ \hline
Milwaukee Wisconsin Oakland California	 & 	Huntsville Alabama Oakland California	 & 	Houston Texas Oakland California	 \\
Shreveport Louisiana Oakland California	 & 	Baltimore Maryland Oakland California	 & 	Chicago Illinois Oakland California	 \\
Irvine California Shreveport Louisiana	 & 	Irvine California Phoenix Arizona	 & 	Denver Colorado Oakland California	 \\
Irvine California Baltimore Maryland	 & 	Arlington Texas Irvine California	 & 	Philadelphia Pennsylvania Oakland Calif	 \\
Sacramento California Henderson Nevada	 & 	Phoenix Arizona Sacramento California	 & 	Portland Oregon Oakland California	 \\
Sacramento California Orlando Florida	 & 	Huntsville Alabama Sacramento California	 & 	Tulsa Oklahoma Irvine California \\\specialrule{.1em}{.05em}{.05em} 
\end{tabular}}\caption{Examples of analogy tests solved. }\label{tab:solvedtest}
}
\end{table*}
\noindent To further investigate these improvements related to the aforementioned parametrisation we focused on one of the most frequent errors in the analogy test, the word \textit{California}.  As we can see from the list of the analogy test solved (Table \ref{tab:solvedtest}) different parametrizations are helpful to solve different types of analogies. For example an increase in the dimensionality increases the accuracy, but mainly in analogy test with words that have a representation in the training data related to a wider set of contexts  (\textit{Houston}:\textit{Texas}; \textit{Chicago}:\textit{Illinois}). The best parametrisation is obtained increasing the negative sampling. As we can see from the examples provided, the analogies are resolved thanks to a contextual similarity between the two pairs (\textit{Huntsville}:\textit{Alabama}; \textit{Oakland}:\textit{California}). In these cases the negative sampling could help to filter out from each representation those words that are not expected to be relevant for the words embeddings.
%\begin{table}[htb]
%\centering
%\resizebox{0.49\textwidth}{!}{%
%\begin{tabular}{l | l | l}
%\textbf{dim = 500 \& n = 10}	 & 	\textbf{\textit{solo} n = 10} 	 & 	\textbf{\textit{solo} dim = 500} 	 \\ \hline
%Milwaukee Wisconsin Oakland California	 & 	Huntsville Alabama Oakland California	 & 	Houston Texas Oakland California	 \\
%Shreveport Louisiana Oakland California	 & 	Baltimore Maryland Oakland California	 & 	Chicago Illinois Oakland California	 \\
%Irvine California Shreveport Louisiana	 & 	Irvine California Phoenix Arizona	 & 	Denver Colorado Oakland California	 \\
%Irvine California Baltimore Maryland	 & 	Arlington Texas Irvine California	 & 	Philadelphia Pennsylvania Oakland Calif	 \\
%Sacramento California Henderson Nevada	 & 	Phoenix Arizona Sacramento California	 & 	Portland Oregon Oakland California	 \\
%Sacramento California Orlando Florida	 & 	Huntsville Alabama Sacramento California	 & 	Tulsa Oklahoma Irvine California	 \\
%\end{tabular}}\caption{Examples of analogy tests solved. }\label{tab:solvedtest}
%\end{table}
%

%As we have seen that the words problematic for all parametrizations are plurals, feminine, currencies, superlatives and ambiguous word we choose to focus on an example that could account for the peculiar problem of some of the aforementioned categories. In the Table \ref{tab:solvedtest2} we present a list of the analogy test solved when is present the word

Similar types of improvement are noticed on analogy tests that contain a challenging word \textit{predire} (\textit{predict}). The results of this analysis are presented in Table \ref{tab:solvedtest2} where it is possible to see that an higher dimensionality improves the accuracy of analogical tests containing open domain verbs (e.g.: \textit{descrivere}, \textit{vedere}). Similarly to the previous case, an higher dimensionality allows for fine grained partitions improving the correct associations between terms. However, also in in this case, the best parametrizations are obtained increasing the negative sampling or both the parameters. As we can see here both the present participle and the past tense pairs are correctly solved. These example provide a preliminary evidence of how negative sampling, filtering out non informative words from the relevant context of each word, is able to build representation by opposition that are beneficial both for semantic and syntactic associations.

\begin{table*}[htb]
\parbox[t]{.31\linewidth}{
\centering
\resizebox{0.31\textwidth}{!}{%
\begin{tabular}{l@{\hskip 0.32in}l@{\hskip 0.25in}l@{\hskip 0.25in}l}\specialrule{.1em}{.05em}{.05em} 
pilotesse & migliore & colori & meloni \\
pere & matrigna & figliastra & sua \\
real & lev & yen & mamma \\
kwanza & vantaggiosissimo & urlano & stimano  \\
aquila & eroina & programmato & impossibilmente \\\specialrule{.1em}{.05em}{.05em} 
\end{tabular}}\caption{Always wrong}\label{tab:missw1}}
\hfill
\parbox[t]{.68\linewidth}{
\centering
\resizebox{0.68\textwidth}{!}{%
\begin{tabular}{l | l | l}\specialrule{.1em}{.05em}{.05em}
\textbf{dim = 500 \& n = 10}	 & 	\textbf{\textit{solo} n = 10} 	 & 	\textbf{\textit{solo} dim = 500} 	 \\ \hline
dire detto predire predetto	 & 	cantare cantato predire predetto	 & 	descrivere descritto predire predetto	 \\
mescolare mescolando predire predicendo	 & 	correre correndo predire predicendo	 & 	vedere visto predire predetto	 \\
predire predicendo generare generando	 & 	generare generando predire predicendo	 & 		 \\
rallentare rallentando predire predicendo	 & 	predire predicendo programmare programmando	 & 		 \\
scoprire scoprendo predire predicendo	 & 	scrivere scrivendo predire predicendo	 & 		 \\\specialrule{.1em}{.05em}{.05em}
\end{tabular}}\caption{Examples of analogy tests solved. }\label{tab:solvedtest2}
}
\end{table*}
\begin{table}[htb]
\centering
\resizebox{0.49\textwidth}{!}{%
\begin{tabular}{l c | l c | l c}\specialrule{.1em}{.05em}{.05em}
SG-200-w5-m5-n10 & \# & SG-500 w5-m5-n1 & \# & SG-500 w5-m5-n10 & \# \\\hline
 capre & 26 & groenlandia & 27 & ratti & 26 \\
 rapidamente & 26 & silenziosamente & 27 & ovviamente & 25 \\
 dolcissimo & 26 & caldissimo & 27 & incredibilmente & 25 \\
 apparentemente & 26 & occhi & 27 & grandissimo & 25 \\
 andato & 26 & greco & 27 & malvolentieri & 25 \\\specialrule{.1em}{.05em}{.05em}
 %washington & 113 & washington & 114 & washington & 111 \\
 %nevada & 77 & nevada & 77 & nevada & 76 \\
 %georgia & 74 & georgia & 77 & georgia & 71 \\
 %pennsylvania & 73 & pennsylvania & 69 & pennsylvania & 62 \\
 %kentucky & 59 & kentucky & 67 & corona & 58
\end{tabular}}\caption{Almost always wrong} \label{tab:mproblem2}
\end{table}
%
%\noindent As we can see from these lists the errors are done on the same words and this because they are the most common in the dataset. We have 217 queries that expect Florida as answer compared to the 55 of Italia. If we go down to the lists we can see that almost always the words in table \ref{tab:mproblem2} are not recovered correctly.
%
Examples of words that almost always are not recovered correctly are presented in Table \ref{tab:mproblem2}. A selected list of words problematic for all parametrizations is shown in Table \ref{tab:missw1}. It contains plurals, feminine, currencies, superlatives and ambiguous words. The low performances on these cases can be explained by the poor coverage of these categories in the training data. In particular, it would be interesting to study the case of feminine and to analyze if it is due to a gender bias in the Italian Wikipedia, as a preliminary analysis of the most frequent errors that persist in all the parametrization seems to suggest.
The words that have been benefited by the increase of $n$ are:
\begin{multicols}{4}
\begin{itemize}[leftmargin=*]\setlength\itemsep{0.3em}
    {\tiny
\item[] ghana
\item[] pakistan
\item[] irlandese
\item[] migliorano
\item[] scrivendo
\item[] slovenia
\item[] giocando
\item[] serbia
\item[] implementano
\item[] ucraino
\item[] zimbabwe
\item[] namibia
\item[] suonano
\item[] maltese
\item[] portoghese
\item[] contessa
\item[] messicano
\item[] giordania \vfill}                         
\end{itemize}
\end{multicols}
%
%
%\begin{table}[!htbp]
%\centering
%\resizebox{0.45\textwidth}{!}{%
%\begin{tabular}{l@{\hskip 0.25in}l@{\hskip 0.25in}l@{\hskip 0.25in}l}
%ghana & pakistan & irlandese & migliorano \\
%scrivendo & slovenia & giocando & serbia \\
%implementano & ucraino & zimbabwe & namibia \\
%suonano & maltese & portoghese & contessa \\
%messicano & giordania \\
%\end{tabular}}\label{tab:missw2}
%\end{table}
%
\noindent the errors that have been introduced increasing this parameter are related to the words in Table \ref{tab:missw3}.
%
%    \begin{multicols}{3}
%    \begin{itemize}[leftmargin=*]\setlength\itemsep{0.3em}
%    {\tiny
%        \item[] irlanda
%        \item[] afghanistan
%        \item[] albania
%        \item[] egiziano
%        \item[] olandese
%        \item[] provvedono
%        \item[] francese
%        \item[] svizzero   }  
%    \end{itemize}
%    \end{multicols}
%
\begin{table}[htb]
\centering
\resizebox{0.45\textwidth}{!}{%
\begin{tabular}{l@{\hskip 0.25in}l@{\hskip 0.25in}l@{\hskip 0.25in}l}\specialrule{.1em}{.05em}{.05em}
irlanda & afghanistan & albania & egiziano \\
olandese & provvedono & francese & svizzero \\\specialrule{.1em}{.05em}{.05em}
\end{tabular}}\caption{New errors}\label{tab:missw3}
\end{table}
It is interesting to notice that given an error in an analogy test, it is possible to find the correct answer in the top five most similar words to the query. Precisely we observed this phenomenon in $26 \%$ of the cases for SG-200-w5-m5-n10, in $27 \%$ of the cases for SG-500-w5-m5-n1 and in $25 \%$ for SG-500-w5-m5-n1. Furthermore, approximately in $50 \%$ of these cases the correct answer is the second most similar.
%
%\subsubsection{Vocabulary issues}
Most of the recovery errors are due to vocabulary issues. In fact, many words of the test set have no correspondence in the developed embedding spaces. This is due to the low frequency of many words that are not in the training corpus or that have been removed from the vocabulary because of their (low) frequency. For this reason we kept the $m$ hyper-parameter very low (e.g., 1 and 5), in counter-tendency with recent works that use larger corpora and then remove infrequent words setting $m$ with high  values (e.g., 50 or 100). In fact, with increasing value of $m$ the number of not given answers increases rapidly. It passes from 300 ($m=1$) to 893 ($m=5$). 

Some of the words that are not present in the vocabulary with $m=1$ include plural verbs (1st person), that probably are not used by a typical Wikipedia editor and remote past verbs (1st person), a tense that in recent years is disappearing from written and spoken Italian. Some of these verbs are:
\begin{multicols}{3}
\begin{itemize}[leftmargin=*]\setlength\itemsep{0.1em}
{\tiny
\item[] giochiamo
\item[] affiliamo
\item[] rallentiamo
\item[] zappiamo
\item[] implementai
\item[] rallentai
\item[] mescolai
\item[] nuotai\vfill}                                 
\end{itemize}
\end{multicols}
%
%\begin{table}[htb]
%\centering
%\resizebox{0.45\textwidth}{!}{%
%\begin{tabular}{l@{\hskip 0.25in}l@{\hskip 0.25in}l@{\hskip 0.25in}l}
% giochiamo & affiliamo & rallentiamo & zappiamo \\
% implementai & rallentai & mescolai & nuotai \\
%\end{tabular}}\label{tab:missw}
%\end{table}
%
In \newcite{berardi2015word} the number of not given answer is 1.220. The accuracy of their embeddings, obtained using a larger corpus and using the hyper-parameters that perform well on English language, is always lower than those obtained with our setting, in both the morphosyntactic and the semantic tasks. This confirms that the regularization of the parameters is crucial for good representation of the embeddings, since the \newcite{berardi2015word}'s model has been trained on a much larger corpus and for this should outperform ours. Furthermore, this model seems to have some tokenization problem.
%In fact, its vocabulary contains entries like:
%\begin{table}[htb]
%\centering
%\resizebox{0.45\textwidth}{!}{%
%\begin{tabular}{l@{\hskip 0.5in}l@{\hskip 0.5in}l}
% giocatori."fonte"collegamenti & industriale.viene & tardi.sempre  \\
% disco.ancora & campionato.tuttavia & televisione.fra  \\ 
%\end{tabular}}\label{tab:missw}
%\end{table}

\section{Conclusions}
We have tested two word representation methods: SG and CBOW training them only on a dump of the Italian Wikipedia. We compared the results of the two models using 12 combinations of hyper-parameters.

We have adopted a simple word analogy test to evaluate the generated word embeddings. The results have shown that increasing the number of dimensions and the number of negative examples improve the performance of both the models.

These types of improvement seems to be beneficial only for the semantic relationships. On the contrary the syntactical relationship are negatively affected by the low frequency of many of its terms. This should be related to the morphological complexity of Italian. In the future it would be helpful to represent the spatial relationship regarding specific syntactical domain in order to evaluate the contribution of  hyper-parametrization to syntactical relationship accuracy. Moreover future work will include the testing of these word embedding parametrizations in practical applications (e.g. analysis of patents'descriptions and books' corpora).

\footnotesize
\subsubsection*{Acknowledgments}
Part of this work has been conducted during a collaboration of the first author with DocFlow Italia. All the experiments in this paper have been conducted on the SCSCF multiprocessor cluster system at University Ca' Foscari of Venice.

% include your own bib file like this:
\bibliography{bib}

\begin{thebibliography}{}

\bibitem[\protect\citename{Al-Rfou \bgroup et al.\egroup }2013]{al2013polyglot}
Rami Al-Rfou, Bryan Perozzi, and Steven Skiena.
\newblock 2013.
\newblock Polyglot: Distributed word representations for multilingual nlp.
\newblock {\em CoNLL-2013}, page 183.

\bibitem[\protect\citename{Attardi and Simi}2014]{attardi2014dependency}
Giuseppe Attardi and Maria Simi.
\newblock 2014.
\newblock Dependency parsing techniques for information extraction.

\bibitem[\protect\citename{Attardi \bgroup et al.\egroup
  }2014]{attardi2014adapting}
Giuseppe Attardi, Vittoria Cozza, and Daniele Sartiano.
\newblock 2014.
\newblock Adapting linguistic tools for the analysis of italian medical
  records.

\bibitem[\protect\citename{Attardi \bgroup et al.\egroup
  }2016]{attardi2016convolutional}
Giuseppe Attardi, Daniele Sartiano, Chiara Alzetta, Federica Semplici, and
  Largo~B Pontecorvo.
\newblock 2016.
\newblock Convolutional neural networks for sentiment analysis on italian
  tweets.
\newblock In {\em CLiC-it/EVALITA}.

\bibitem[\protect\citename{Bengio \bgroup et al.\egroup
  }2003]{bengio2003neural}
Yoshua Bengio, R{\'e}jean Ducharme, Pascal Vincent, and Christian Jauvin.
\newblock 2003.
\newblock A neural probabilistic language model.
\newblock {\em Journal of machine learning research}, 3(Feb):1137--1155.

\bibitem[\protect\citename{Berardi \bgroup et al.\egroup
  }2015]{berardi2015word}
Giacomo Berardi, Andrea Esuli, and Diego Marcheggiani.
\newblock 2015.
\newblock Word embeddings go to italy: A comparison of models and training
  datasets.
\newblock In {\em IIR}.

\bibitem[\protect\citename{Collobert and Weston}2008]{collobert2008unified}
Ronan Collobert and Jason Weston.
\newblock 2008.
\newblock A unified architecture for natural language processing: Deep neural
  networks with multitask learning.
\newblock In {\em Proceedings of the 25th international conference on Machine
  learning}, pages 160--167. ACM.

\bibitem[\protect\citename{Firth}1935]{firth1935technique}
John~Rupert Firth.
\newblock 1935.
\newblock The technique of semantics.
\newblock {\em Transactions of the philological society}, 34(1):36--73.

\bibitem[\protect\citename{Goldberg}2017]{goldberg2017neural}
Yoav Goldberg.
\newblock 2017.
\newblock Neural network methods for natural language processing.
\newblock {\em Synthesis Lectures on Human Language Technologies},
  10(1):1--309.

\bibitem[\protect\citename{Harris}1954]{harris1954distributional}
Zellig~S Harris.
\newblock 1954.
\newblock Distributional structure. word, 10 (2-3): 146--162. reprinted in
  fodor, j. a and katz, jj (eds.), readings in the philosophy of language.

\bibitem[\protect\citename{Kim \bgroup et al.\egroup }2016]{kim2016character}
Yoon Kim, Yacine Jernite, David Sontag, and Alexander~M Rush.
\newblock 2016.
\newblock Character-aware neural language models.
\newblock In {\em AAAI}, pages 2741--2749.

\bibitem[\protect\citename{Levy \bgroup et al.\egroup
  }2014]{levy2014linguistic}
Omer Levy, Yoav Goldberg, and Israel Ramat-Gan.
\newblock 2014.
\newblock Linguistic regularities in sparse and explicit word representations.
\newblock In {\em CoNLL}, pages 171--180.

\bibitem[\protect\citename{Levy \bgroup et al.\egroup }2015]{levy2015improving}
Omer Levy, Yoav Goldberg, and Ido Dagan.
\newblock 2015.
\newblock Improving distributional similarity with lessons learned from word
  embeddings.
\newblock {\em Transactions of the Association for Computational Linguistics},
  3:211--225.

\bibitem[\protect\citename{Mikolov \bgroup et al.\egroup
  }2010]{mikolov2010recurrent}
Tomas Mikolov, Martin Karafi{\'a}t, Lukas Burget, Jan Cernock{\`y}, and Sanjeev
  Khudanpur.
\newblock 2010.
\newblock Recurrent neural network based language model.
\newblock In {\em Interspeech}, volume~2, page~3.

\bibitem[\protect\citename{Mikolov \bgroup et al.\egroup
  }2013]{mikolov2013efficient}
Tomas Mikolov, Kai Chen, Greg Corrado, and Jeffrey Dean.
\newblock 2013.
\newblock Efficient estimation of word representations in vector space.
\newblock {\em arXiv preprint arXiv:1301.3781}.

\bibitem[\protect\citename{Stemle}2016]{stemle2016bot}
Egon~W Stemle.
\newblock 2016.
\newblock bot. zen@ evalita 2016-a minimally-deep learning pos-tagger (trained
  for italian tweets).
\newblock In {\em CLiC-it/EVALITA}.

\bibitem[\protect\citename{Tamburini}2016]{tamburini2016bilstm}
Fabio Tamburini.
\newblock 2016.
\newblock A bilstm-crf pos-tagger for italian tweets using morphological
  information.
\newblock In {\em CLiC-it/EVALITA}.

\end{thebibliography}
\bibliographystyle{acl}

\end{document}